\documentclass{article}

\usepackage{iclr2017_conference,times}
\iclrfinalcopy
\PassOptionsToPackage{numbers,comma,square}{natbib}

\usepackage[utf8]{inputenc} 
\usepackage[T1]{fontenc}    
\usepackage[bookmarks=false]{hyperref}       
\usepackage{url}            
\usepackage{booktabs}       
\usepackage{amsfonts}       
\usepackage{nicefrac}       
\usepackage{microtype}      

\usepackage{graphicx}
\usepackage{caption}
\usepackage{url}
\usepackage{subfigure}
\usepackage{amsmath}
\usepackage{amssymb}
\usepackage{colortbl}
\usepackage{booktabs}
\usepackage{stmaryrd}
\usepackage{float}
\usepackage{dsfont} 
\usepackage{upgreek} 
\usepackage[raggedright]{titlesec}
\usepackage{tabularx}
\usepackage{lipsum}

\bibpunct{[}{]}{,}{n}{,}{,}
\usepackage{adjustbox}
\usepackage{wrapfig}

\usepackage{xcolor}
\definecolor{darkgreen}{rgb}{0,0.6,0}

\newcommand{\mb}{\mathbf}

\DeclareMathOperator{\diag} {diag}

\DeclareMathOperator*{\argmax}{argmax}

\renewcommand{\cite}[1]{\citep{#1}}

\newcounter{lastromanpg}

\def\Id{{\mathbb{I}}}

\def\X{{\mathcal{X}}}

\def\Z{{\mathcal{Z}}}

\title{Density estimation using Real NVP}

\author{
  Laurent Dinh\thanks{Work was done when author was at Google Brain.} \\
  Montreal Institute for Learning Algorithms\\
  University of Montreal\\
  Montreal, QC H3T1J4\\
  \AND
  Jascha Sohl-Dickstein\\
  Google Brain \\
  \And
  Samy Bengio\\
  Google Brain \\
}

\begin{document}

\maketitle

\begin{abstract}
Unsupervised learning of probabilistic models is a central yet challenging problem in machine learning.
  Specifically, designing models with tractable learning, sampling,
  inference and evaluation is crucial in solving this task. We extend the space
  of such models using real-valued non-volume preserving (real NVP)
  transformations, a set of powerful, stably invertible, and learnable transformations,
  resulting in an unsupervised learning algorithm with exact log-likelihood computation,
  exact and efficient sampling, exact and efficient inference of latent variables,
  and an interpretable latent space. We
  demonstrate its ability to model natural images on four datasets through
  sampling, log-likelihood evaluation, and latent variable
  manipulations.
\end{abstract}
\section{Introduction}
The domain of representation learning has undergone tremendous advances due to improved supervised learning techniques. However, unsupervised learning has the potential to leverage large pools of unlabeled data, and extend these advances to modalities that are otherwise impractical or impossible.

One principled approach to unsupervised learning is generative probabilistic modeling. Not only do generative probabilistic models have the ability to create novel content, they also have a wide range of reconstruction related applications including inpainting \citep{theis2015generative, oord2016pixel, DBLP:conf/icml/Sohl-DicksteinW15}, denoising \citep{balle2015density}, colorization \citep{zhang2016colorful}, and super-resolution \citep{bruna2015super}.

As data of interest are generally high-dimensional and highly structured, the challenge in this domain is building models that are powerful enough to capture its complexity yet still trainable.
We address this challenge by introducing \emph{real-valued non-volume preserving (real NVP) transformations}, a tractable yet expressive approach to modeling high-dimensional data.

This model can perform {\em efficient and exact} inference, sampling and log-density estimation of data points. Moreover, the architecture presented in this paper enables {\em exact and efficient} reconstruction of input images from the hierarchical features extracted by this model.
%


\section{Related work}
\begin{figure}
\begin{center}
{\renewcommand{\arraystretch}{2}
\begin{tabular}{m{1.25in}ccc}
& Data space $\X$ & & Latent space $\Z$
\\
\parbox{1.25in}{
\textbf{Inference}
$\begin{aligned}
\qquad x &\sim \hat{p}_{X} &\\
\qquad z &= f\left(x\right)
\end{aligned}$
}
&
\raisebox{-.43\height}{
\includegraphics[width=1.5in]{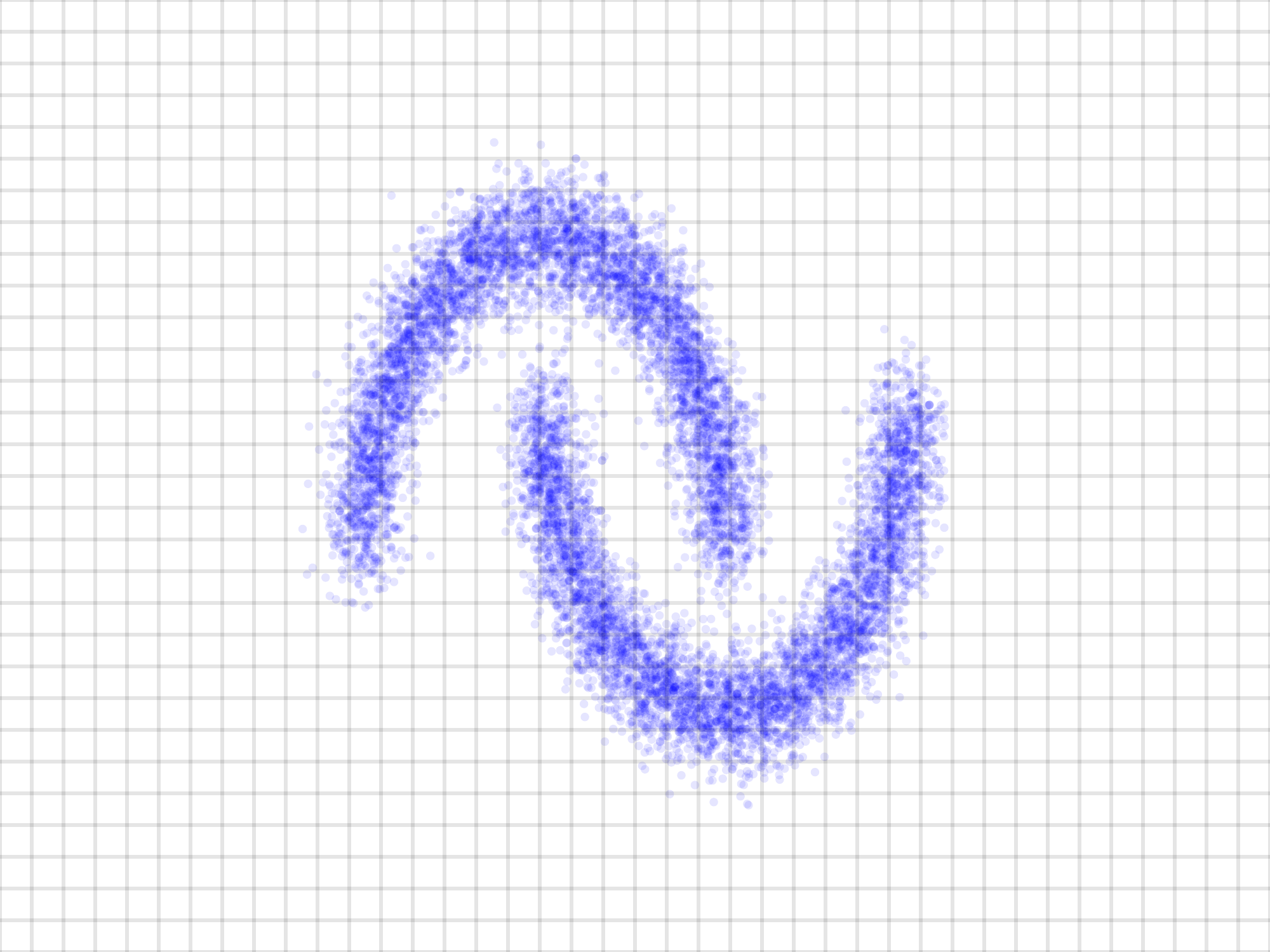}
}
&
\parbox{.25in}{
$\begin{aligned}
\Rightarrow
\end{aligned}$
}
&
\raisebox{-.43\height}{
\includegraphics[width=1.5in]{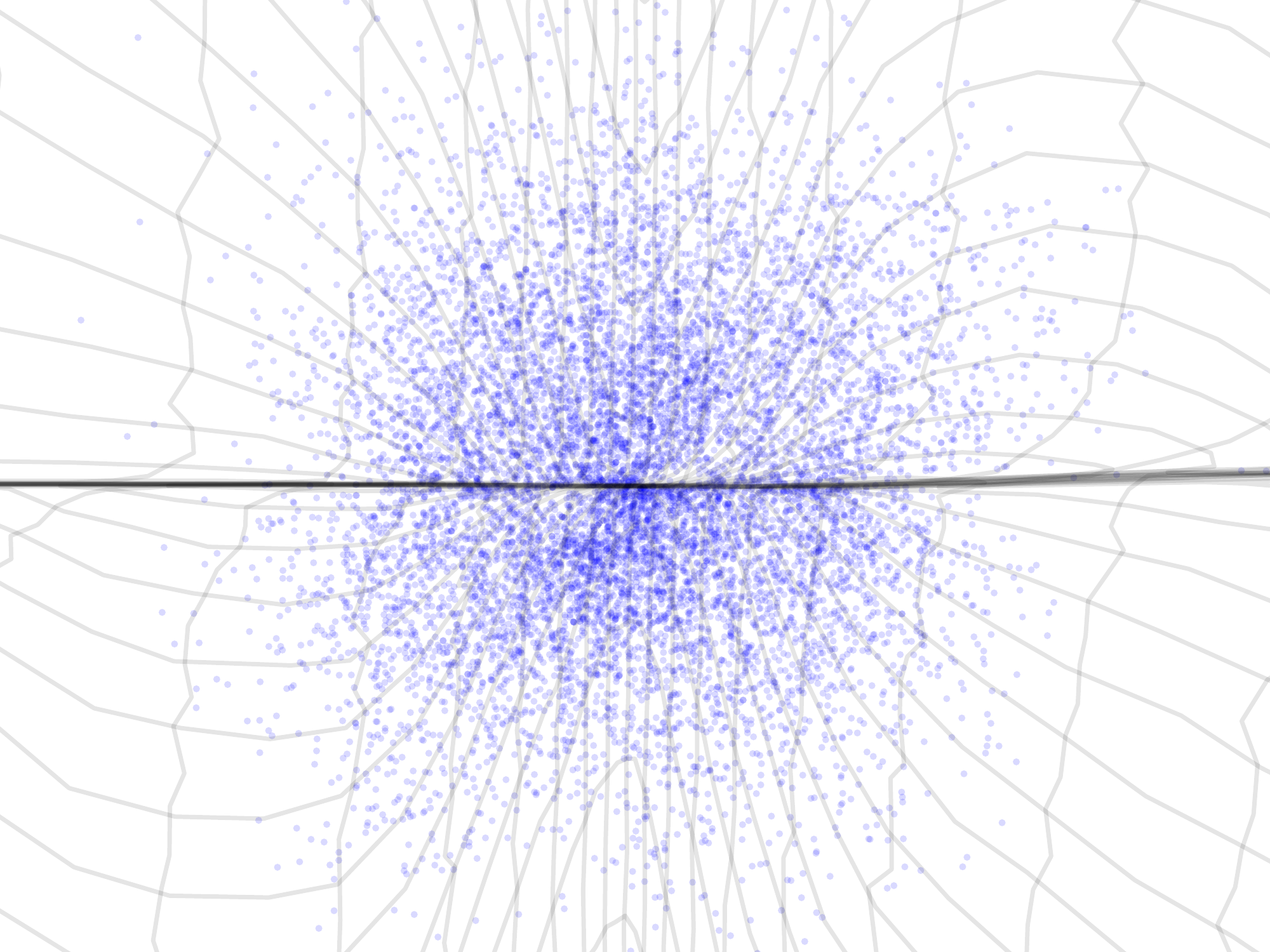}
}
\\[0.5in] 
\parbox{1.25in}{
\textbf{Generation}
$\begin{aligned}
\qquad z &\sim p_{Z} &
\\
\qquad x &= f^{-1}\left(z\right)
\end{aligned}$
}
 &
\raisebox{-.43\height}{
\includegraphics[width=1.5in]{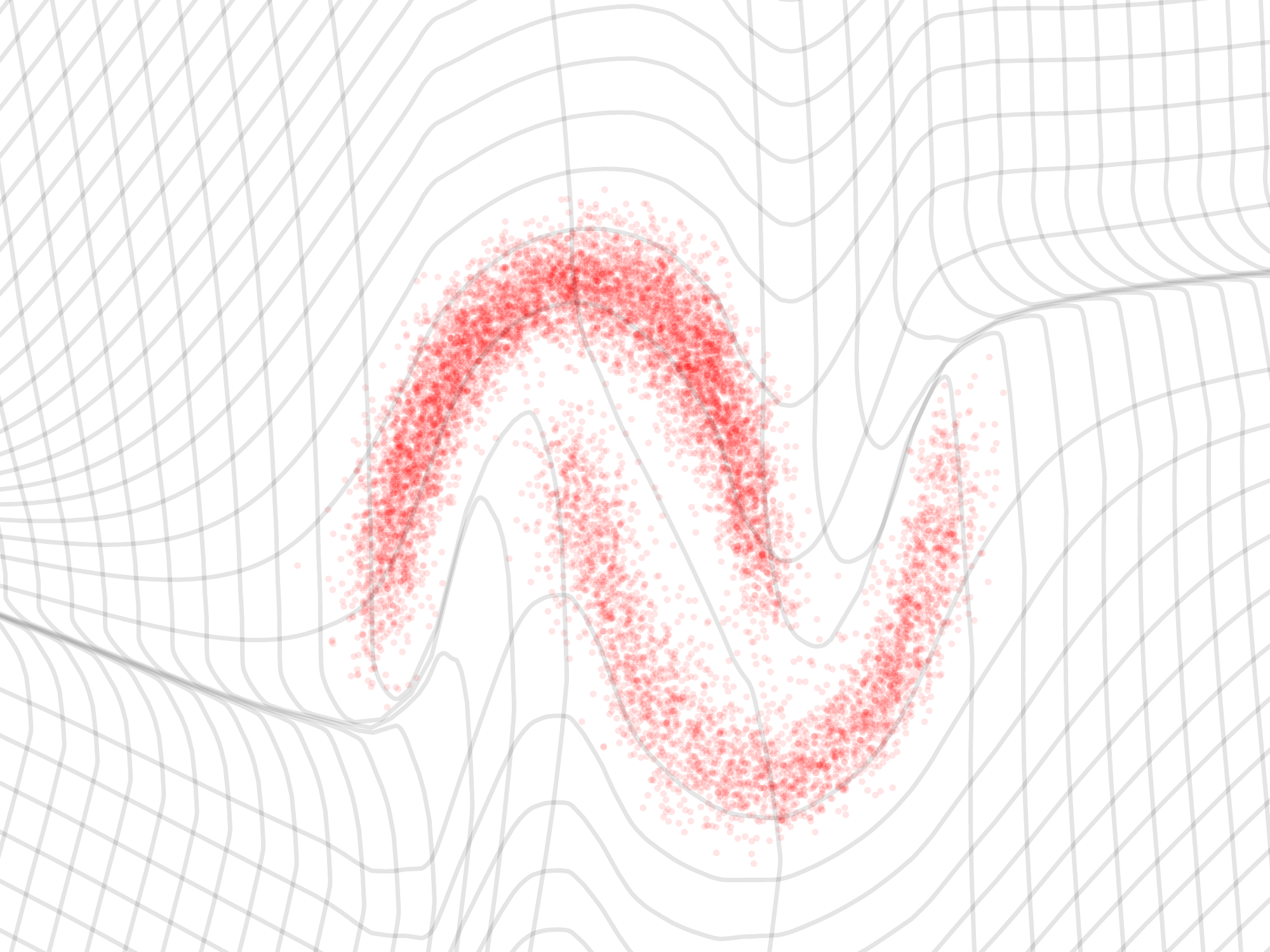}
}
&
\parbox{.25in}{
$\begin{aligned}
\Leftarrow
\end{aligned}$
}
&
\raisebox{-.43\height}{
\includegraphics[width=1.5in]{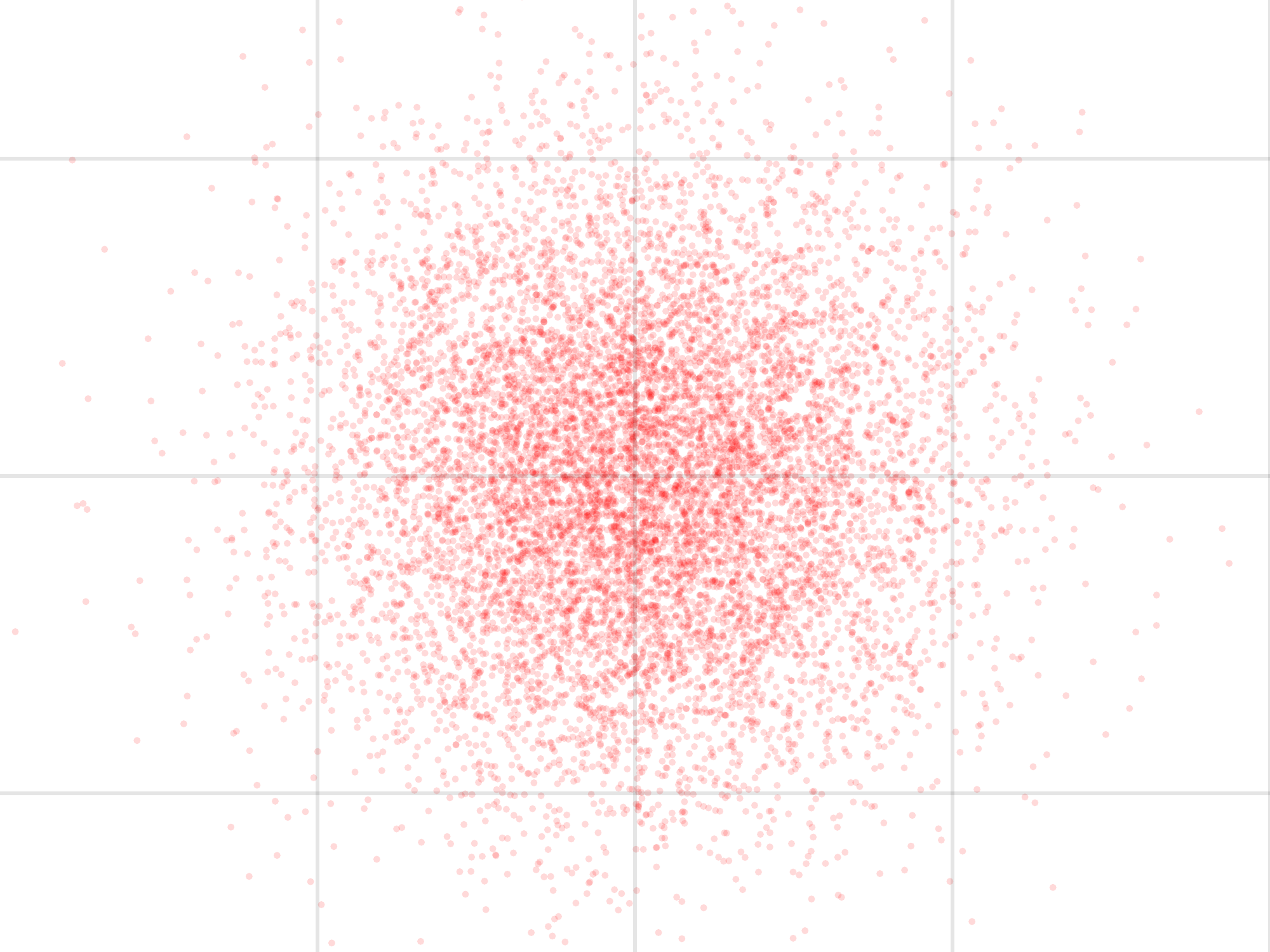}
}
\end{tabular}}
\end{center}
    \caption{Real NVP learns an invertible, stable, mapping between a data distribution $\hat{p}_{X}$ and a latent distribution $p_{Z}$ (typically a Gaussian).
    Here we show a mapping that has been learned on a toy 2-d dataset.
    The function $f\left(x\right)$ maps samples $x$ from the data distribution in the upper left into approximate samples $z$ from the latent distribution, in the upper right. This corresponds to exact inference of the latent state given the data.
    The inverse function, $f^{-1}\left(z\right)$, maps samples $z$ from the latent distribution in the lower right into approximate samples $x$ from the data distribution in the lower left. This corresponds to exact generation of samples from the model.
    The transformation of grid lines in $\X$ and $\Z$ space is additionally illustrated for both $f\left(x\right)$ and $f^{-1}\left(z\right)$.
    }
    \label{fig:spaghetti}
\end{figure}
Substantial work on probabilistic generative models has focused on training models using maximum likelihood.
One class of maximum likelihood models are those described by \emph{probabilistic undirected graphs}, such as \emph{Restricted Boltzmann Machines} \citep{smolensky1986information} and \emph{Deep Boltzmann Machines} \citep{salakhutdinov2009deep}. These models are trained by taking advantage of the conditional independence property of their bipartite structure to allow efficient exact or approximate posterior inference on latent variables. However, because of the intractability of the associated marginal distribution over latent variables, their training, evaluation, and sampling procedures necessitate the use of approximations like \emph{Mean Field inference} and \emph{Markov Chain Monte Carlo}, whose convergence time for such complex models remains undetermined, often resulting in generation of highly correlated samples. Furthermore, these approximations can often hinder their performance \citep{berglund2013stochastic}. 

\emph{Directed graphical models} are instead defined in terms of an \emph{ancestral sampling procedure}, which is appealing both for its conceptual and computational simplicity. They lack, however, the conditional independence structure of undirected models, making exact and approximate posterior inference on latent variables cumbersome \citep{saul1996mean}. Recent advances in \emph{stochastic variational inference} \citep{hoffman2013stochastic} and \emph{amortized inference} \citep{dayan1995helmholtz, mnih2014neural, kingma2013auto, rezende2014stochastic}, allowed efficient approximate inference and learning of deep directed graphical models by maximizing a variational lower bound on the log-likelihood \citep{neal1998view}. In particular, the \emph{variational autoencoder algorithm} \citep{kingma2013auto, rezende2014stochastic} simultaneously learns a \emph{generative network}, that maps gaussian latent variables $z$ to samples $x$, and a matched \emph{approximate inference network} that maps samples $x$ to a semantically meaningful latent representation $z$, by exploiting the \emph{reparametrization trick} \citep{williams1992simple}.
Its success in leveraging recent advances in backpropagation \citep{rumelhart1988learning, lecun2012efficient} in deep neural networks resulted in its adoption for several applications ranging from speech synthesis \citep{chung2015recurrent} to language modeling \citep{bowman2015generating}. Still, the approximation in the inference process limits its ability to learn high dimensional deep representations, motivating recent work in improving approximate inference \citep{maaloe2016auxiliary, rezende2015variational, salimans2014markov, tran2015variational, burda2015importance,  DBLP:conf/icml/Sohl-DicksteinW15, kingma2016improving}.

Such approximations can be avoided altogether by abstaining from using latent variables. \emph{Auto-regressive models} \citep{frey1998graphical, bengio1999modeling, larochelle2011neural, DBLP:journals/corr/GermainGML15} can implement this strategy while typically retaining a great deal of flexibility. This class of algorithms tractably models the joint distribution by decomposing it into a product of conditionals using the \emph{probability chain rule} according to a fixed ordering over dimensions, simplifying log-likelihood evaluation and sampling. Recent work in this line of research has taken advantage of recent advances in \emph{recurrent networks} \citep{rumelhart1988learning}, in particular \emph{long-short term memory} \citep{DBLP:journals/neco/HochreiterS97}, and \emph{residual networks} \citep{DBLP:journals/corr/HeZR016, DBLP:journals/corr/HeZRS15} in order to learn state-of-the-art generative image models \citep{theis2015generative, oord2016pixel} and language models \citep{DBLP:journals/corr/JozefowiczVSSW16}.
The ordering of the dimensions, although often arbitrary, can be critical to the training of the model \citep{vinyals2015order}.
The sequential nature of this model limits its computational efficiency. For example, its sampling procedure is sequential and non-parallelizable, which can become cumbersome in applications like speech and music synthesis, or real-time rendering..
Additionally, there is no natural latent representation associated with autoregressive models, and they have not yet been shown to be useful for semi-supervised learning.

\emph{Generative Adversarial Networks} (GANs) \citep{DBLP:conf/nips/GoodfellowPMXWOCB14} on the other hand can train any differentiable generative network by avoiding the maximum likelihood principle altogether. Instead, the generative network is associated with a \emph{discriminator network} whose task is to distinguish between samples and real data. Rather than using an intractable log-likelihood, this discriminator network provides the training signal in an adversarial fashion. Successfully trained GAN models \citep{DBLP:conf/nips/GoodfellowPMXWOCB14, DBLP:conf/nips/DentonCSF15, DBLP:journals/corr/RadfordMC15} can consistently generate sharp and realistically looking samples \citep{DBLP:journals/corr/LarsenSW15}. However, metrics that measure the diversity in the generated samples are currently intractable \citep{DBLP:journals/corr/TheisOB15, gregor2016towards, im2016generating}. Additionally, instability in their training process \citep{DBLP:journals/corr/RadfordMC15} requires careful hyperparameter tuning to avoid diverging behavior.

Training such a generative network $g$ that maps latent variable $z \sim p_{Z}$ to a sample $x \sim p_{X}$ does not in theory require a discriminator network as in GANs, or approximate inference as in variational autoencoders. Indeed, if $g$ is bijective, it can be trained through maximum likelihood using the \emph{change of variable formula}:
\begin{align}
p_{X}(x) = p_{Z}(z) \left\vert \det\left(\frac{\partial g(z)}{\partial z^T}\right)\right\vert^{-1}
.
\end{align}

This formula has been discussed in several papers including the maximum likelihood formulation of \emph{independent components analysis} (ICA) \citep{bell1995information, hyvarinen2004independent}, gaussianization \citep{NIPS1994_901, chen2000gaussianization} and deep density models \citep{bengio1991artificial, rippel2013high, dinh2014nice, balle2015density}.
As the existence proof of nonlinear ICA solutions \citep{hyvarinen1999nonlinear} suggests, auto-regressive models can be seen as tractable instance of maximum likelihood nonlinear ICA, where the residual corresponds to the independent components.
However, naive application of the change of variable formula produces models which are computationally expensive and poorly conditioned, and so large scale models of this type have not entered general use.

\section{Model definition}
In this paper, we will tackle the problem of learning highly nonlinear models in high-dimensional continuous spaces through maximum likelihood. In order to optimize the log-likelihood, we introduce a more flexible class of architectures that enables the computation of log-likelihood on continuous data using the change of variable formula. Building on our previous work in \citep{dinh2014nice}, we define a powerful class of bijective functions which enable exact and tractable density evaluation and exact and tractable inference.
Moreover, the resulting cost function does not to rely on a fixed form reconstruction cost such as square error \citep{DBLP:journals/corr/LarsenSW15, DBLP:journals/corr/RadfordMC15}, and generates sharper samples as a result. Also, this flexibility helps us leverage recent advances in batch normalization \citep{ioffe2015batch} and residual networks \citep{DBLP:journals/corr/HeZRS15, DBLP:journals/corr/HeZR016} to define a very deep multi-scale architecture with multiple levels of abstraction.

\subsection{Change of variable formula}
Given an observed data variable $x \in X$,
a simple prior probability distribution $p_{Z}$ on a latent variable $z \in Z$,
and a bijection $f: X \rightarrow Z$ (with $g = f^{-1}$),
the change of variable formula defines a model distribution on $X$ by
\begin{align}
p_{X}(x) &= p_{Z}\big(f(x)\big) \left|\det\left(\cfrac{\partial f(x)}{\partial x^T} \right)\right|
\label{eq:change-variables}\\
\log\left(p_{X}(x)\right) &= \log\Big(p_{Z}\big(f(x)\big)\Big) + \log\left(\left|\det\left(\frac{\partial f(x)}{\partial x^T}\right)\right|\right)
,
\end{align}
where $\frac{\partial f(x)}{\partial x^T}$ is the Jacobian of $f$ at $x$.

Exact samples from the resulting distribution can be generated by using the inverse transform sampling rule \citep{devroye1986sample}. A sample $z \sim p_{Z}$ is drawn in the latent space, and its inverse image $x = f^{-1}(z) = g(z)$ generates a sample in the original space. Computing the density on a point $x$ is accomplished by computing the density of its image $f(x)$ and multiplying by the associated Jacobian determinant $\det\left(\frac{\partial f(x)}{\partial x^T}\right)$. See also Figure \ref{fig:spaghetti}. Exact and efficient inference enables the accurate and fast evaluation of the model.


\subsection{Coupling layers}
\begin{figure}
    \centering
    \subfigure[Forward propagation]{
      \includegraphics[width=.2\textwidth]{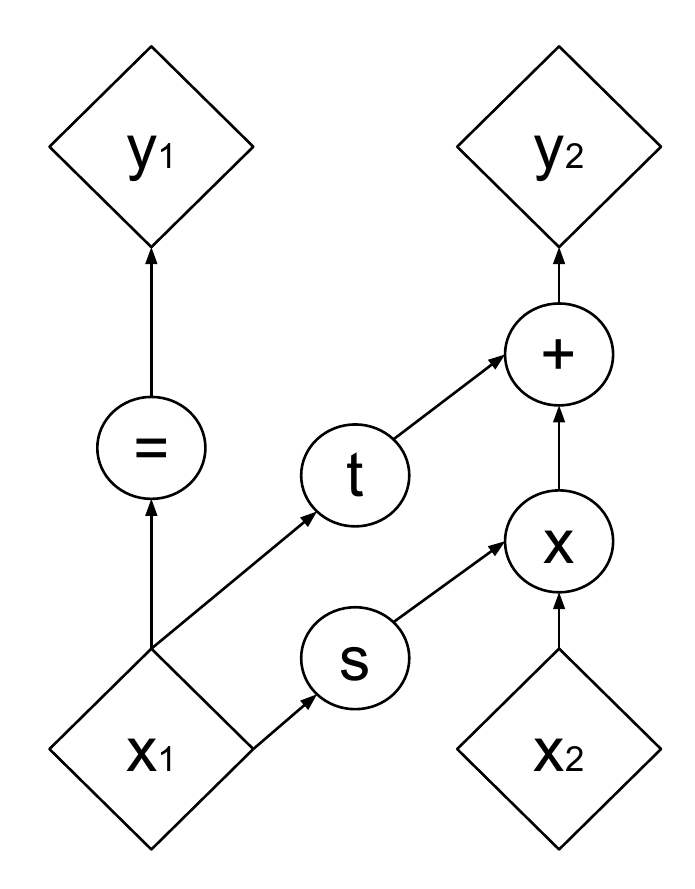}
      \qquad
      \label{fig:coupling}} ~~~~~~
    \subfigure[Inverse propagation
    ]{
    	\qquad
      \includegraphics[width=.2\textwidth]{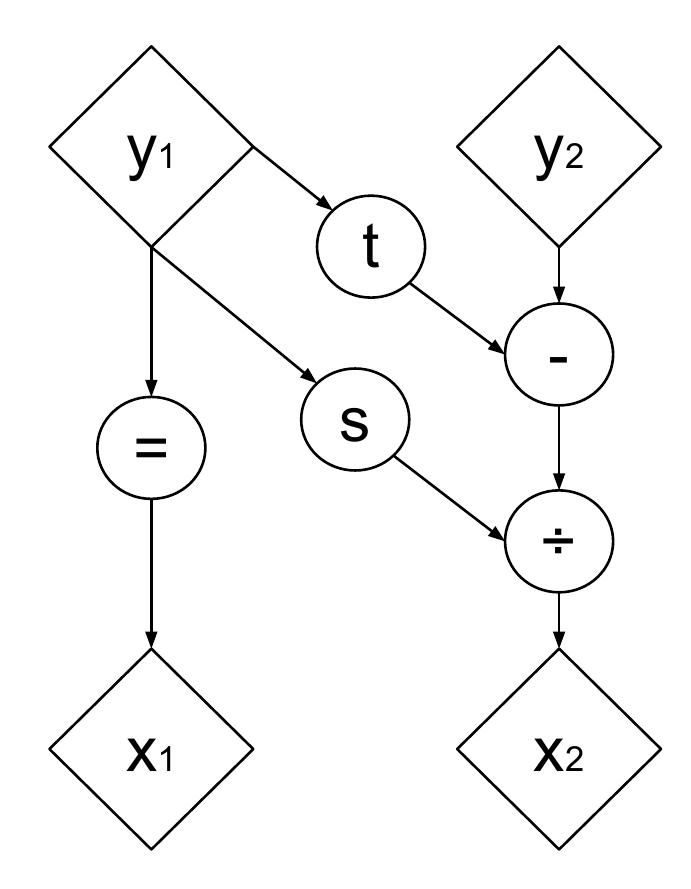}
      \label{fig:inverse_coupling}}
    \caption{Computational graphs for forward and inverse propagation. A coupling layer applies a simple invertible transformation consisting of scaling followed by addition of a constant offset to one part $\mb x_{2}$ of the input vector conditioned on the remaining part of the input vector $\mb x_{1}$. Because of its simple nature, this transformation is both easily invertible and possesses a tractable determinant.
    However, the conditional nature of this transformation, captured by the functions $s$ and $t$, significantly increase the flexibility of this otherwise weak function.
	The forward and inverse propagation operations have identical computational cost. }
\end{figure}

Computing the Jacobian of functions with high-dimensional domain and codomain and computing the determinants of large matrices are in general computationally very expensive. This combined with the restriction to bijective functions makes Equation \ref{eq:change-variables} appear impractical for modeling arbitrary distributions.

As shown however in \citep{dinh2014nice}, by careful design of the function $f$, a bijective model can be learned which is both tractable and extremely flexible.
As computing the Jacobian determinant of the transformation is crucial to effectively train using this principle, this work exploits the simple observation that \emph{the determinant of a triangular matrix can be efficiently computed} as the product of its diagonal terms.

We will build a flexible and tractable bijective function by stacking a sequence of simple bijections.
In each simple bijection,
part of the input vector is updated using a function which is simple to invert,
but which depends on the remainder of the input vector in a complex way.
We refer to each of these simple bijections as an \emph{affine coupling layer}.
Given a $D$ dimensional input $x$ and $d < D$, the output $y$ of an affine coupling layer follows the equations
\begin{align}
y_{1:d} &= x_{1:d}\\
y_{d+1:D} &= x_{d+1:D} \odot \exp\big(s(x_{1:d})\big) + t(x_{1:d})
,
\end{align}
where $s$ and $t$ stand for scale and translation, and are functions from $R^{d} \mapsto R^{D - d}$, and $\odot$ is the Hadamard product or element-wise product (see Figure \ref{fig:coupling}).

\subsection{Properties}
The Jacobian of this transformation is
\begin{align}
\frac{\partial y}{\partial x^T} = \left[\begin{array}{cc}
\Id_{d} & 0 \\
\frac{\partial y_{d+1:D}}{\partial x_{1:d}^T} & \diag\big(\exp\left[s\left(x_{1:d}\right)\right]\big)
\end{array} \right]
,
\end{align}
where $\diag\big(\exp\left[s\left(x_{1:d}\right)\right]\big)$ is the diagonal matrix whose diagonal elements correspond to the vector $\exp\left[s\left(x_{1:d}\right)\right]$. Given the observation that this Jacobian is triangular, we can efficiently compute its determinant as $\exp\left[\sum_{j}{s\left(x_{1:d}\right)_j}\right]$. Since computing the Jacobian determinant of the coupling layer operation does not involve computing the Jacobian of $s$ or $t$, those functions can be arbitrarily complex. We will make them deep convolutional neural networks. Note that the hidden layers of $s$ and $t$ can have more features than their input and output layers.

Another interesting property of these coupling layers in the context of defining probabilistic models is their invertibility. Indeed, computing the inverse is no more complex than the forward propagation (see Figure \ref{fig:inverse_coupling}),
\begin{align}
\begin{cases}
y_{1:d} &= x_{1:d} \\
y_{d+1:D} &= x_{d+1:D} \odot \exp\big(s(x_{1:d})\big) + t(x_{1:d})
\end{cases}\\
\Leftrightarrow
\begin{cases}
x_{1:d} &= y_{1:d} \\
x_{d+1:D} &= \big(y_{d+1:D} - t(y_{1:d})\big) \odot \exp\big(-s(y_{1:d})\big),
\end{cases}
\end{align}
meaning that sampling is as efficient as inference for this model.
Note again that computing the inverse of the coupling layer does not require computing the inverse of $s$ or $t$, so these functions can be arbitrarily complex and difficult to invert.

\subsection{Masked convolution}
\begin{figure}
    \centering \includegraphics[width=.6\textwidth]{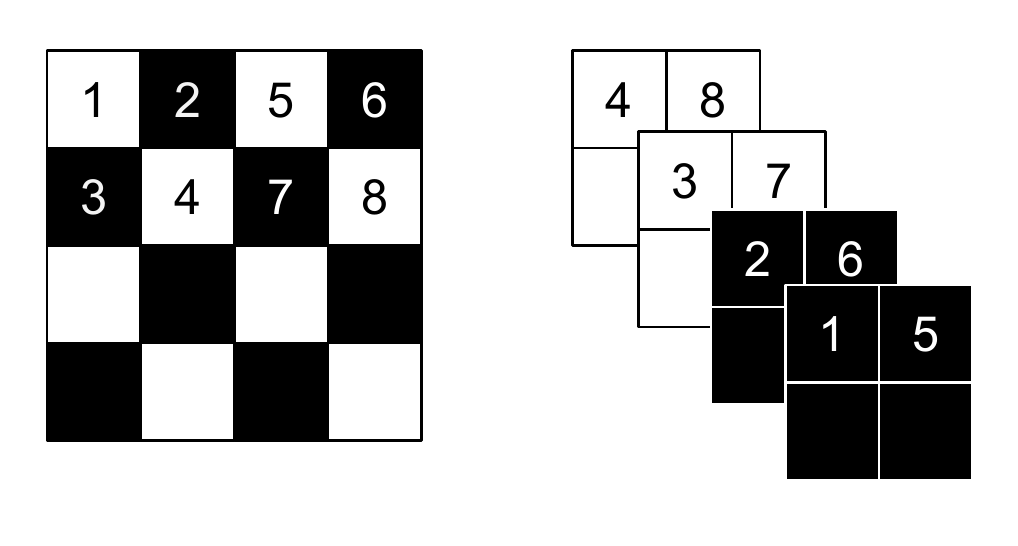}
\vspace{-10pt}
    \caption{Masking schemes for affine coupling layers. On the left, a spatial checkerboard pattern mask. On the right, a channel-wise masking. The squeezing operation reduces the $4 \times 4 \times 1$ tensor (on the left) into a $2 \times 2 \times 4$ tensor (on the right). Before the squeezing operation, a checkerboard pattern is used for coupling layers while a channel-wise masking pattern is used afterward. 
    }
    \label{fig:squeezing}
\end{figure}

Partitioning can be implemented using a binary mask $b$, and using the functional form for $y$,
\begin{align}
y = b \odot x + (1 - b) \odot \Big(x \odot \exp\big(s(b \odot x)\big) + t(b \odot x)\Big)
.
\end{align}
We use two partitionings that exploit the local correlation structure of images: spatial checkerboard patterns, and channel-wise masking (see Figure \ref{fig:squeezing}). The spatial checkerboard pattern mask has value $1$ where the sum of spatial coordinates is odd, and $0$ otherwise. The channel-wise mask $b$ is $1$ for the first half of the channel dimensions and $0$ for the second half.
For the models presented here, both $s(\cdot)$ and $t(\cdot)$ are rectified convolutional networks.

\subsection{Combining coupling layers}
\begin{figure}[h]
    \centering \subfigure[In this alternating pattern, units which remain identical in one transformation are modified in the next.]{
      \includegraphics[width=.6\textwidth]{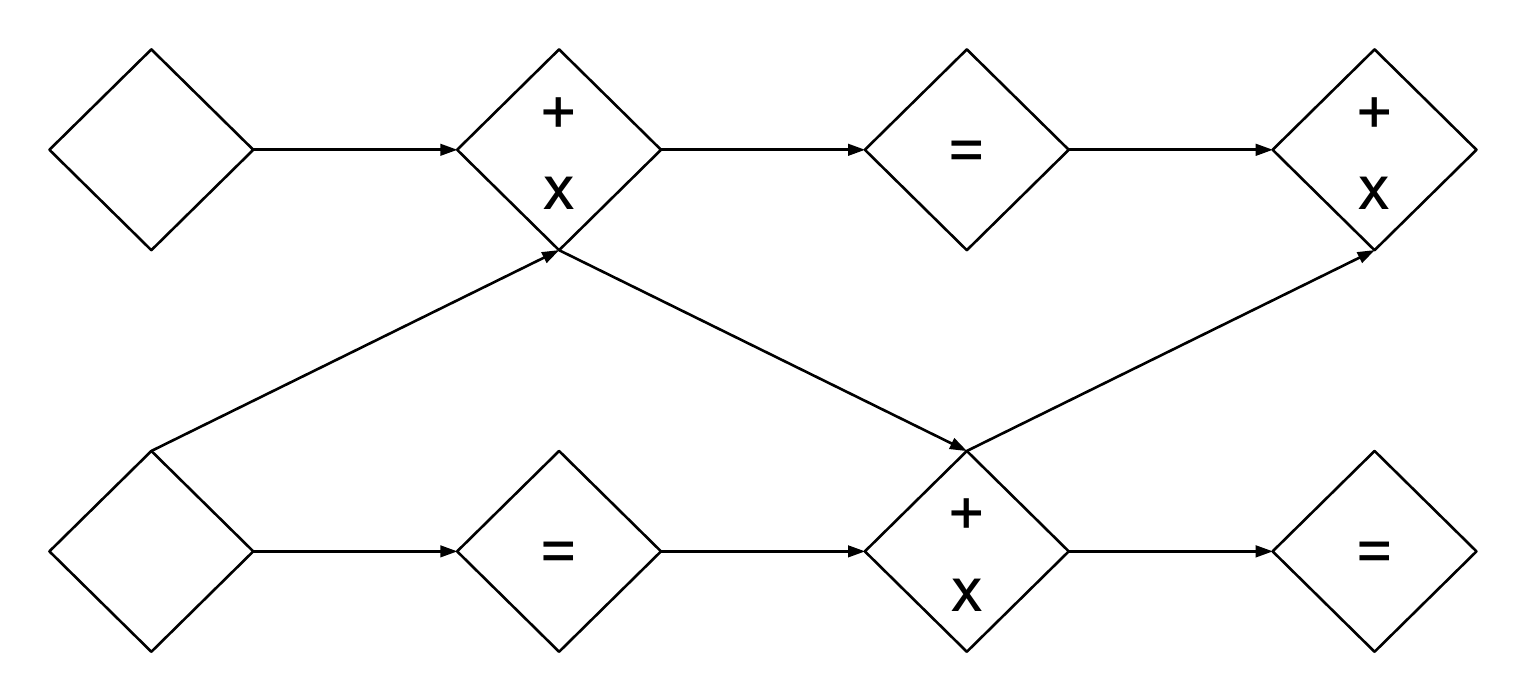}
      \label{fig:composition}} ~~~~~~
    \subfigure[Factoring out variables. At each step, half the variables are directly modeled as Gaussians, while the other half undergo further transformation.]{
      \includegraphics[width=.25\textwidth]{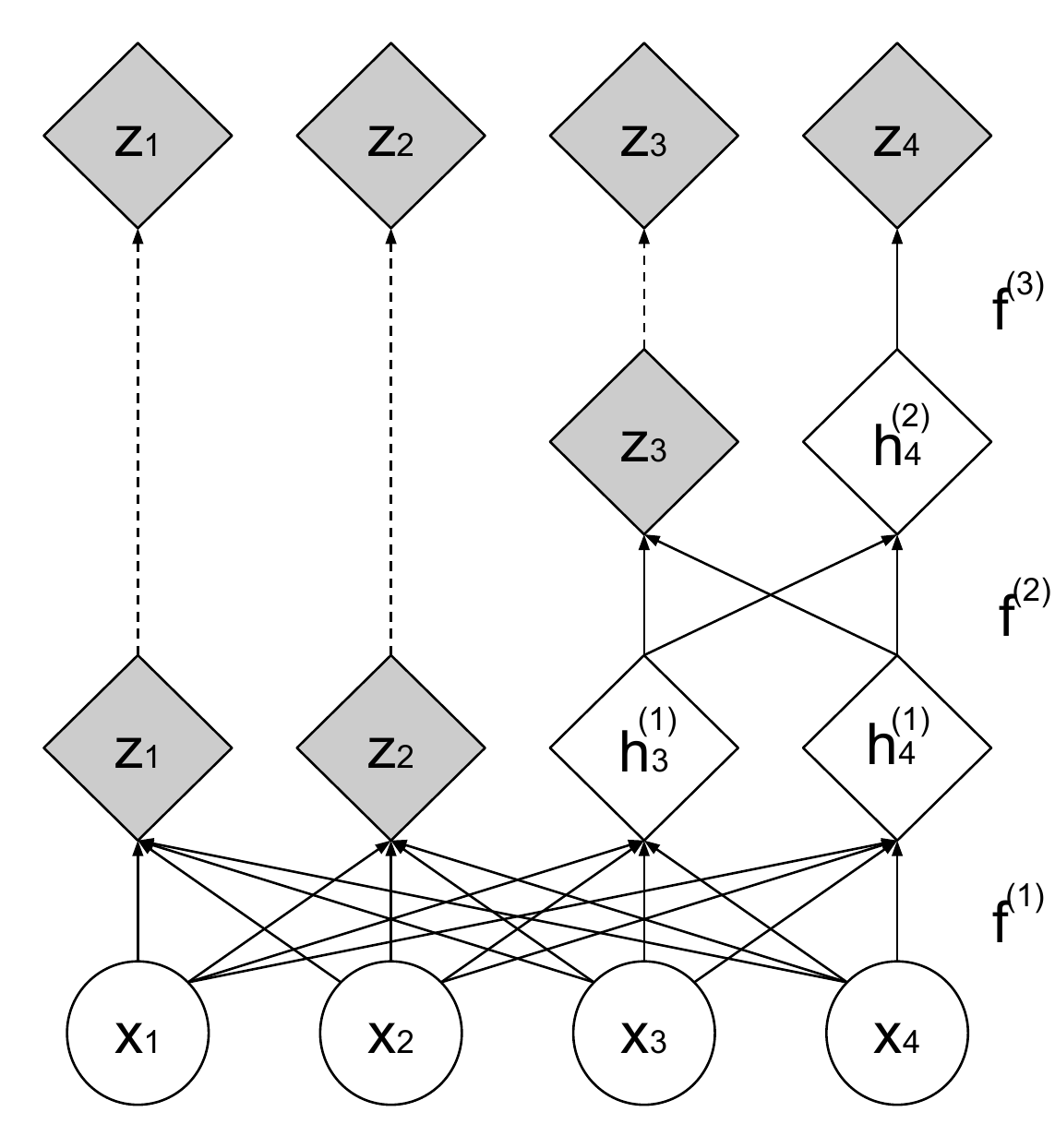}
      \label{fig:factor_out}}
    \caption{Composition schemes for affine coupling layers.}
\end{figure}

Although coupling layers can be powerful, their forward transformation leaves some components unchanged.
This difficulty can be overcome by composing coupling layers in an alternating pattern, such that the components that are left unchanged in one coupling layer are updated in the next (see Figure \ref{fig:composition}).

The Jacobian determinant of the resulting function remains tractable, relying on the fact that
\begin{align}
\cfrac{\partial (f_{b} \circ f_{a})}{\partial x_{a}^{T}}(x_{a}) &= \cfrac{\partial f_{a}}{\partial x_{a}^{T}}(x_{a}) \cdot \cfrac{\partial f_{b}}{\partial x_{b}^{T}}\big(x_{b}=f_{a}(x_{a})\big) \\
\det(A \cdot B) &= \det(A) \det(B).
\end{align}
Similarly, its inverse can be computed easily as
\begin{align}
(f_{b} \circ f_{a})^{-1} = f_{a}^{-1} \circ f_{b}^{-1}.
\end{align}


\subsection{Multi-scale architecture}
\label{section:multiscale}
We implement a multi-scale architecture using a squeezing operation: for each channel, it divides the image into subsquares of shape $2 \times 2 \times c$, then reshapes them into subsquares of shape $1 \times 1 \times 4c$. The squeezing operation transforms an $s \times s \times c$ tensor into an $\frac{s}{2} \times \frac{s}{2} \times 4c$ tensor (see Figure  \ref{fig:squeezing}), effectively trading spatial size for number of channels.

At each scale, we combine several operations into a sequence: we first apply three coupling layers with alternating checkerboard masks, then perform a squeezing operation, and finally apply three more coupling layers with alternating channel-wise masking. The channel-wise masking is chosen so that the resulting partitioning is not redundant with the previous checkerboard masking (see Figure \ref{fig:squeezing}). For the final scale, we only apply four coupling layers with alternating checkerboard masks. 

Propagating a $D$ dimensional vector through all the coupling layers would be cumbersome, in terms of computational and memory cost, and in terms of the number of parameters that would need to be trained.
For this reason we follow the design choice of \citep{simonyan2014very} and factor out half of the dimensions at regular intervals (see Equation \ref{eq:factor_out}).
We can define this operation recursively (see Figure \ref{fig:factor_out}),
\begin{align}
h^{(0)} &= x \\
(z^{(i + 1)}, h^{(i + 1)}) &= f^{(i + 1)}(h^{(i)})
\label{eq:factor_out} \\
z^{(L)} &= f^{(L)}(h^{(L - 1)})
\label{eq:final_factor}\\
z &= (z^{(1)}, \dots, z^{(L)})
\label{eq:concat_factors}
.
\end{align}
In our experiments, we use this operation for $i < L$.
The sequence of coupling-squeezing-coupling operations described above is performed per layer when computing $f^{(i)}$ (Equation \ref{eq:factor_out}).
At each layer, as the spatial resolution is reduced, the number of hidden layer features in $s$ and $t$ is doubled.
All variables which have been factored out at different scales are concatenated to obtain the final transformed output (Equation \ref{eq:concat_factors}).

As a consequence, the model must Gaussianize units which are factored out at a finer scale (in an earlier layer) before those which are factored out at a coarser scale (in a later layer). This results in the definition of intermediary levels of representation \citep{salakhutdinov2009deep, rezende2014stochastic} corresponding to more local, fine-grained features as shown in Appendix \ref{app:latent}.

Moreover, Gaussianizing and factoring out units in earlier layers has the practical benefit of distributing the loss function throughout the network, following the philosophy similar to guiding intermediate layers using intermediate classifiers \citep{lee2014deeply}. It also reduces significantly the amount of computation and memory used by the model, allowing us to train larger models.


\subsection{Batch normalization}
To further improve the propagation of training signal, we use deep residual networks \citep{DBLP:journals/corr/HeZRS15, DBLP:journals/corr/HeZR016} with batch normalization \citep{ioffe2015batch} and weight normalization \citep{badrinarayanan2015understanding, salimans2016weight} in $s$ and $t$.
As described in Appendix \ref{sec batch norm} we introduce and use a novel variant of batch normalization which is based on a running average over recent minibatches, and is thus more robust when training with very small minibatches.

We also use apply batch normalization to the whole coupling layer output.
The effects of batch normalization are easily included in the Jacobian computation, since it acts as a linear rescaling on each dimension. That is, given the estimated batch statistics $\tilde{\mu}$ and $\tilde{\sigma}^{2}$, the rescaling function
\begin{align}
x \mapsto \frac{x - \tilde{\mu}}{\sqrt{\tilde{\sigma}^{2} + \epsilon}}
\end{align}
has a Jacobian determinant
\begin{align}
\left(\prod_{i}{(\tilde{\sigma}_{i}^{2} + \epsilon)}\right)^{-\frac{1}{2}}.
\end{align}
This form of batch normalization can be seen as similar to reward normalization in deep reinforcement learning \citep{mnih2015human, van2016learning}.

We found that the use of this technique not only allowed training with a deeper stack of coupling layers, but also alleviated the instability problem that practitioners often encounter when training conditional distributions with a scale parameter through a gradient-based approach.

\section{Experiments}
\subsection{Procedure}
The algorithm described in Equation \ref{eq:change-variables} shows how to learn distributions on unbounded space. In general, the data of interest have bounded magnitude. For examples, the pixel values of an image typically lie in $[0, 256]^{D}$ after application of the recommended jittering procedure \citep{uria2013rnade, DBLP:journals/corr/TheisOB15}. In order to reduce the impact of boundary effects, we instead model the density of $\mbox{logit}(\alpha + (1 - \alpha) \odot \frac{x}{256})$, where $\alpha$ is picked here as $.05$.
We take into account this transformation when computing log-likelihood and bits per dimension. We also augment the CIFAR-10, CelebA and LSUN datasets during training to also include horizontal flips of the training examples. 

We train our model on four natural image datasets: \emph{CIFAR-10} \citep{krizhevsky2009learning}, \emph{Imagenet} \citep{russakovsky2015imagenet}, \emph{Large-scale Scene Understanding (LSUN)} \citep{yu2015construction}, \emph{CelebFaces Attributes (CelebA)} \citep{liu2015faceattributes}. More specifically, we train on the downsampled to $32 \times 32$ and $64 \times64$ versions of Imagenet \citep{oord2016pixel}. For the LSUN dataset, we train on the \emph{bedroom}, \emph{tower} and \emph{church} outdoor categories. The procedure for LSUN is the same as in \citep{DBLP:journals/corr/RadfordMC15}: we downsample the image so that the smallest side is $96$ pixels and take random crops of $64 \times64$. For CelebA, we use the same procedure as in \citep{DBLP:journals/corr/LarsenSW15}: we take an approximately central crop of $148 \times 148$ then resize it to $64 \times 64$.

We use the multi-scale architecture described in Section \ref{section:multiscale} and use deep convolutional residual networks in the coupling layers with rectifier nonlinearity and skip-connections as suggested by \citep{oord2016pixel}. To compute the scaling functions $s$, we use a {\em hyperbolic tangent} function multiplied by a learned scale, whereas the translation function $t$ has an affine output. Our multi-scale architecture is repeated recursively until the input of the last recursion is a $4 \times 4 \times c$ tensor. For datasets of images of size $32 \times 32$, we use $4$ residual blocks with $32$ hidden feature maps for the first coupling layers with checkerboard masking. Only $2$ residual blocks are used for images of size $64 \times 64$. We use a batch size of $64$. For CIFAR-10, we use $8$ residual blocks, $64$ feature maps, and downscale only once. We optimize with ADAM \citep{kingma2014adam} with default hyperparameters and use an $L_{2}$ regularization on the weight scale parameters with coefficient $5 \cdot 10^{-5}$.

We set the prior $p_{Z}$ to be an isotropic unit norm Gaussian. However, any distribution could be used for $p_{Z}$, including distributions that are also learned during training, such as from an auto-regressive model, or (with slight modifications to the training objective) a variational autoencoder.

\subsection{Results}
\begin{table}
    \begin{center}
    \begin{tabular}{| c | c | c | c | c |}
      \hline
      \textbf{Dataset} & \textbf{PixelRNN} \citep{oord2016pixel} & \textbf{Real NVP} & \textbf{Conv DRAW \citep{gregor2016towards}} & \textbf{IAF-VAE \citep{kingma2016improving}} \\ \hline
      \textbf{CIFAR-10} & 3.00 & 3.49 & < 3.59 & < 3.28 \\ \hline
      \textbf{Imagenet ($32 \times 32$)} & 3.86 (3.83) & 4.28 (4.26) &  < 4.40 (4.35) & \cellcolor{black!25} \\ \hline
      \textbf{Imagenet ($64 \times 64$)} & 3.63 (3.57) & 3.98 (3.75) & < 4.10 (4.04) & \cellcolor{black!25} \\ \hline
      \textbf{LSUN (bedroom)} & \cellcolor{black!25} & 2.72 (2.70) & \cellcolor{black!25} & \cellcolor{black!25} \\ \hline
      \textbf{LSUN (tower)} & \cellcolor{black!25} & 2.81 (2.78) & \cellcolor{black!25} & \cellcolor{black!25} \\ \hline
      \textbf{LSUN (church outdoor)} & \cellcolor{black!25} & 3.08 (2.94) & \cellcolor{black!25} & \cellcolor{black!25} \\ \hline
      \textbf{CelebA} & \cellcolor{black!25} & 3.02 (2.97) & \cellcolor{black!25} & \cellcolor{black!25} \\ \hline
    \end{tabular}
    \caption{Bits/dim results for CIFAR-10, Imagenet, LSUN datasets and CelebA. Test results for CIFAR-10 and validation results for Imagenet, LSUN and CelebA (with training results in parenthesis for reference). 
    \label{fig:quant}}
    \end{center}
\end{table}

We show in Table \ref{fig:quant} that the number of bits per dimension, while not improving over the Pixel RNN \citep{oord2016pixel} baseline, is competitive with other generative methods. As we notice that our performance increases with the number of parameters, larger models are likely to further improve performance.
For CelebA and LSUN, the bits per dimension for the validation set was decreasing throughout training, so little overfitting is expected.

\begin{figure}
\begin{center}
    \includegraphics[width=.47\textwidth]{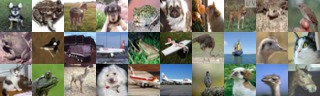} \hfill
    \includegraphics[width=.47\textwidth]{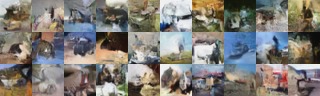} \\\vspace{1mm}

    \includegraphics[width=.47\textwidth]{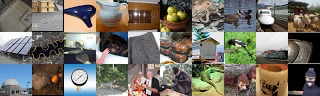} \hfill
    \includegraphics[width=.47\textwidth]{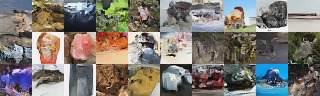} \\\vspace{1mm}

    \includegraphics[width=.47\textwidth]{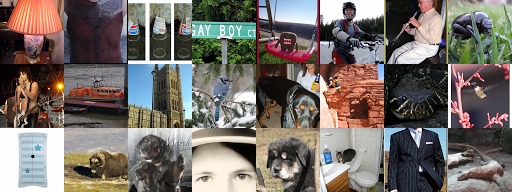} \hfill
    \includegraphics[width=.47\textwidth]{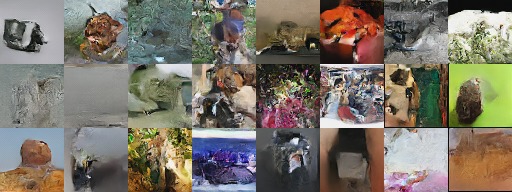} \\\vspace{1mm}
    \includegraphics[width=.47\textwidth]{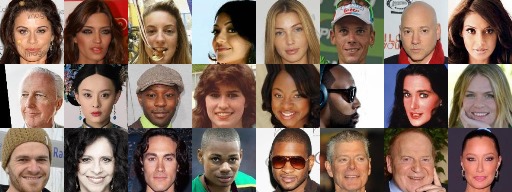} \hfill
    \includegraphics[width=.47\textwidth]{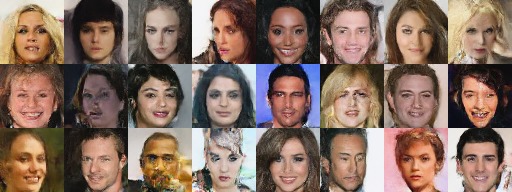} \\\vspace{1mm}
    \includegraphics[width=.47\textwidth]{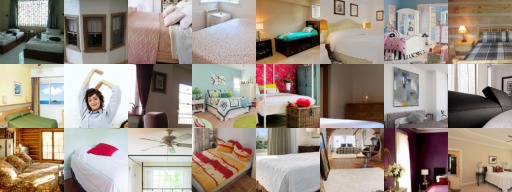} \hfill
    \includegraphics[width=.47\textwidth]{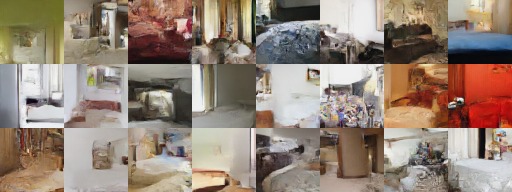} 
    \caption{On the left column, examples from the dataset. On the right column, samples from the model trained on the dataset. The datasets shown in this figure are in order: CIFAR-10, Imagenet ($32 \times 32$), Imagenet ($64 \times 64$), CelebA, LSUN (bedroom).
    }
    \label{fig:samples}
\end{center}
\end{figure}

We show in Figure \ref{fig:samples} samples generated from the model with training examples from the dataset for comparison. As mentioned in \citep{DBLP:journals/corr/TheisOB15, gregor2016towards}, maximum likelihood is a principle that values diversity over sample quality in a limited capacity setting. As a result, our model outputs sometimes highly improbable samples as we can notice especially on CelebA. As opposed to variational autoencoders, the samples generated from our model look not only globally coherent but also sharp. Our hypothesis is that as opposed to these models, real NVP does not rely on fixed form reconstruction cost like an $L_{2}$ norm which tends to reward capturing low frequency components more heavily than high frequency components. Unlike autoregressive models, sampling from our model is done very efficiently as it is parallelized over input dimensions. 
On Imagenet and LSUN, our model seems to have captured well the notion of background/foreground and lighting interactions such as luminosity and consistent light source direction for reflectance and shadows. 

\begin{figure}
\begin{center}
    \includegraphics[width=.47\textwidth]{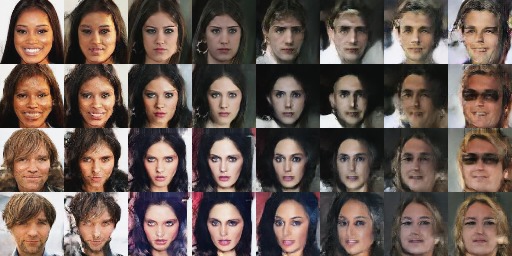} \hfill
    \includegraphics[width=.47\textwidth]{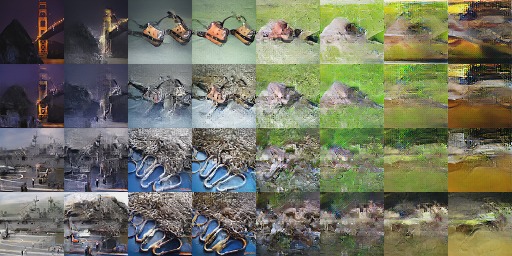} \\\vspace{1mm}
    \includegraphics[width=.47\textwidth]{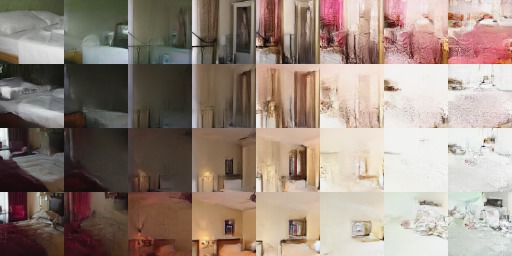} \hfill
    \includegraphics[width=.47\textwidth]{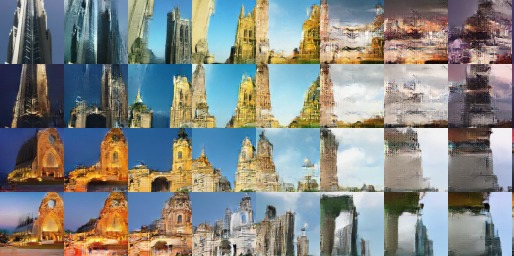}
\end{center}
    \caption{Manifold generated from four examples in the dataset. Clockwise from top left: CelebA, Imagenet ($64 \times 64$), LSUN (tower), LSUN (bedroom).}
    \label{fig:manifold}
\end{figure}

We also illustrate the smooth semantically consistent meaning of our latent variables.
In the latent space, we define a manifold based on four validation examples $z_{(1)}$, $z_{(2)}$, $z_{(3)}$, $z_{(4)}$, and parametrized by two parameters $\phi$ and $\phi'$ by,
\begin{align}
z = \cos(\phi) \left(\cos(\phi') z_{(1)} + \sin(\phi') z_{(2)}\right) + \sin(\phi) \left(\cos(\phi') z_{(3)} + \sin(\phi') z_{(4)}\right)
.
\label{eq:manifold}
\end{align}
We project the resulting manifold back into the data space by computing $g(z)$. Results are shown Figure \ref{fig:manifold}. We observe that the model seems to have organized the latent space with a notion of meaning that goes well beyond pixel space interpolation. More visualization are shown in the Appendix. To further test whether the latent space has a consistent semantic interpretation, we trained a class-conditional model on CelebA, and found that the learned representation had a consistent semantic meaning across class labels (see Appendix \ref{app attr change}).

\section{Discussion and conclusion}
In this paper, we have defined a class of invertible functions with tractable Jacobian determinant, enabling exact and tractable log-likelihood evaluation, inference, and sampling.
We have shown that this class of generative model achieves competitive performances, both in terms of sample quality and log-likelihood.
Many avenues exist to further improve the functional form of the transformations, for instance by exploiting the latest advances in dilated convolutions \citep{yu2015multi} and residual networks architectures \citep{DBLP:journals/corr/TargAL16}.

This paper presented a technique bridging the gap between auto-regressive models, variational autoencoders, and generative adversarial networks. Like auto-regressive models, it allows tractable and exact log-likelihood evaluation for training. It allows however a much more flexible functional form, similar to that in the generative model of variational autoencoders. This allows for fast and exact sampling from the model distribution.
Like GANs, and unlike variational autoencoders, our technique does not require the use of a fixed form reconstruction cost, and instead defines a cost in terms of higher level features, generating sharper images.
Finally, unlike both variational autoencoders and GANs, our technique is able to learn a semantically meaningful latent space which is as high dimensional as the input space. This may make the algorithm particularly well suited to semi-supervised learning tasks, as we hope to explore in future work.

Real NVP generative models can additionally be conditioned on additional variables (for instance class labels) to create a structured output algorithm. More so, as the resulting class of invertible transformations can be treated as a probability distribution in a modular way, it can also be used to improve upon other probabilistic models like auto-regressive models and variational autoencoders. For variational autoencoders, these transformations could be used both to enable a more flexible reconstruction cost  \citep{DBLP:journals/corr/LarsenSW15} and a more flexible stochastic inference distribution \citep{ rezende2015variational}. Probabilistic models in general can also benefit from batch normalization techniques as applied in this paper.

The definition of powerful and trainable invertible functions can also benefit domains other than generative unsupervised learning. For example, in reinforcement learning, these invertible functions can help extend the set of functions for which an $\argmax$ operation is tractable for {\em continuous Q-learning} \citep{gu2016continuous} or find representation where local linear Gaussian approximations are more appropriate \citep{watter2015embed}.

\section{Acknowledgments}
The authors thank the developers of Tensorflow \citep{abadi2016tensorflow}.
We thank Sherry Moore, David Andersen and Jon Shlens for their help in implementing the model.
We thank Aäron van den Oord, Yann Dauphin, Kyle Kastner, Chelsea Finn, Maithra Raghu, David Warde-Farley, Daniel Jiwoong Im and Oriol Vinyals for fruitful discussions.
Finally, we thank Ben Poole, Rafal Jozefowicz and George Dahl for their input on a draft of the paper.

\small
\setlength{\bibsep}{0pt plus 0.12ex}
\bibliographystyle{plain}
\bibliography{local}

\iftrue
\newpage
\appendix
\section{Samples}
~ \\

\begin{figure}[H]
    \centering \includegraphics[width=1.\textwidth]{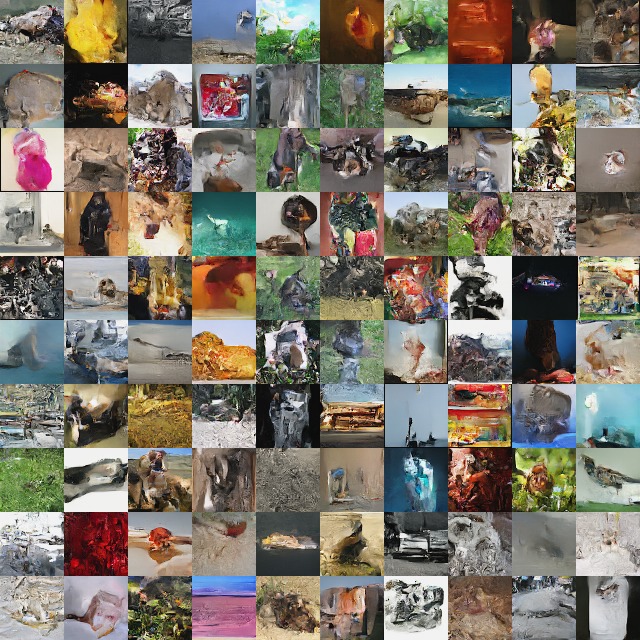}
    \caption{Samples from a model trained on \emph{Imagenet} ($64 \times 64$).}
\vspace{120pt}
\end{figure}

\begin{figure}[H]
\vspace{80pt}
    \centering \includegraphics[width=1.\textwidth]{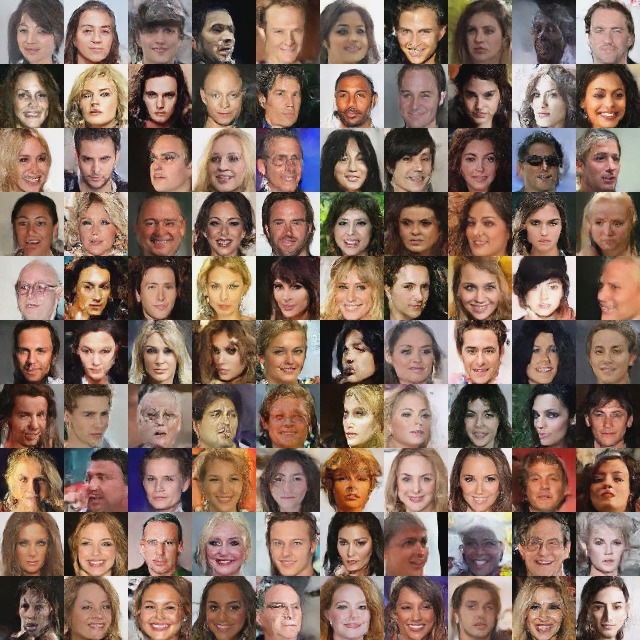}
    \caption{Samples from a model trained on \emph{CelebA}.}
\end{figure}

\begin{figure}[H]
\vspace{80pt}
    \centering \includegraphics[width=1.\textwidth]{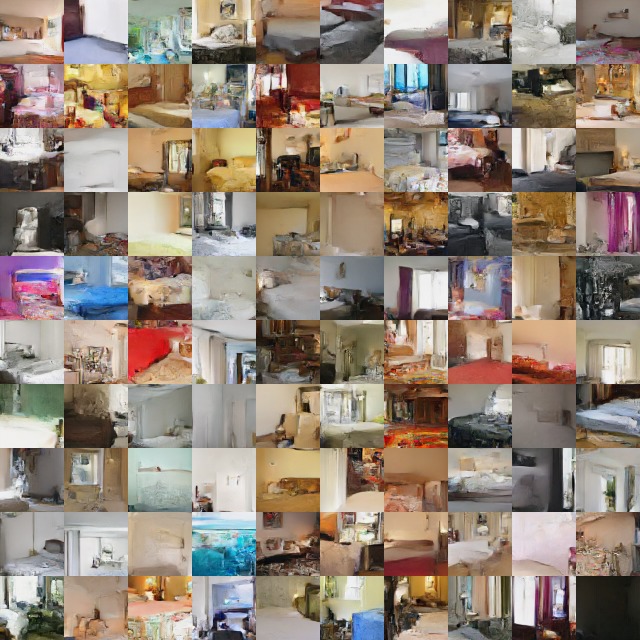}
    \caption{Samples from a model trained on \emph{LSUN} (\emph{bedroom} category).}
\end{figure}

\begin{figure}[H]
\vspace{80pt}
    \centering \includegraphics[width=1.\textwidth]{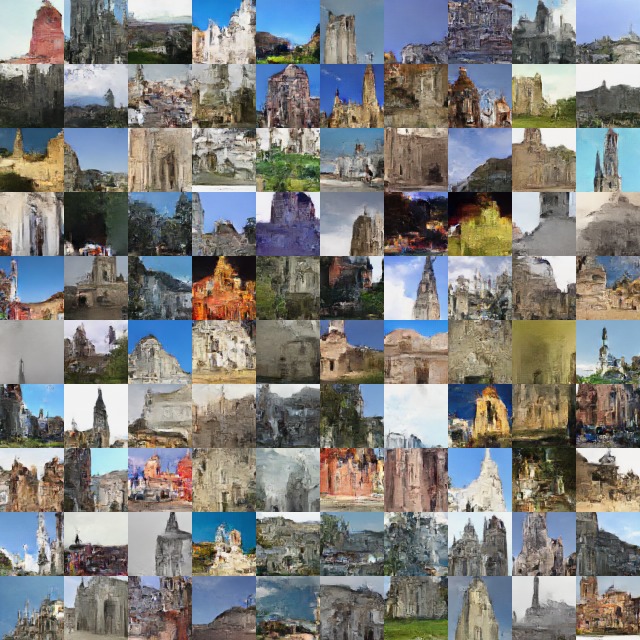}
    \caption{Samples from a model trained on \emph{LSUN} (\emph{church outdoor} category).}
\end{figure}

\begin{figure}[H]
\vspace{80pt}
    \centering \includegraphics[width=1.\textwidth]{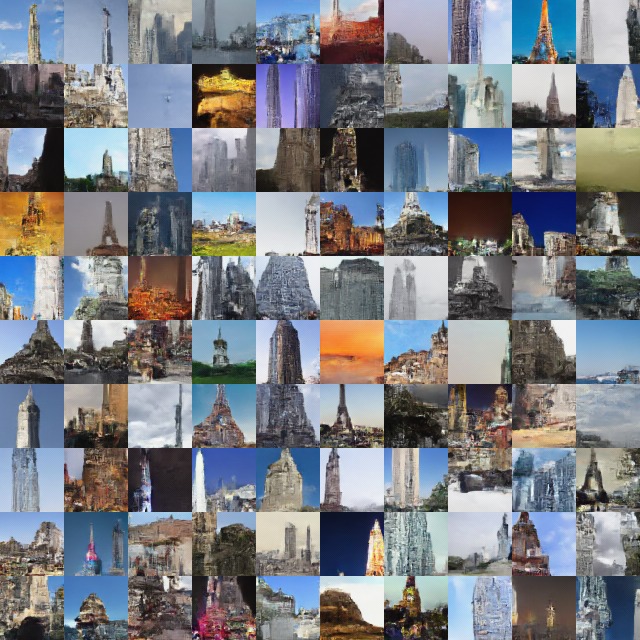}
    \caption{Samples from a model trained on \emph{LSUN} (\emph{tower} category).}
\end{figure}

\newpage

\section{Manifold}
~ \\

\begin{figure}[H]
    \centering \includegraphics[width=1.\textwidth]{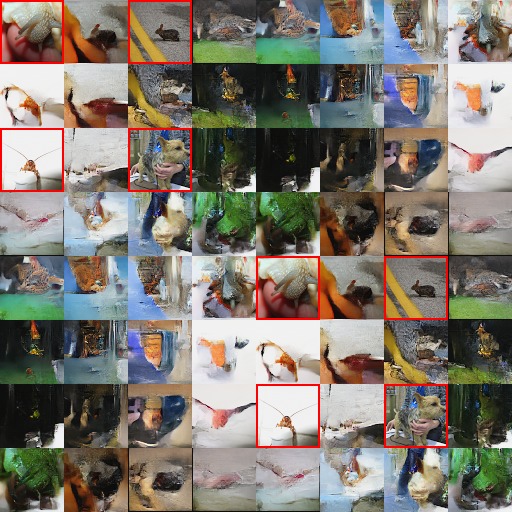}
    \caption{Manifold from a model trained on \emph{Imagenet} ($64 \times 64$). Images with red borders are taken from the validation set, and define the manifold. The manifold was computed as described in Equation \ref{eq:manifold}, where the x-axis corresponds to $\phi$, and the y-axis to $\phi'$, and where $\phi, \phi' \in \{0, \frac{\pi}{4}, \cdots, \frac{7\pi}{4}\}$.
    }
\vspace{80pt}
\end{figure}

\begin{figure}[H]
\vspace{80pt}
    \centering \includegraphics[width=1.\textwidth]{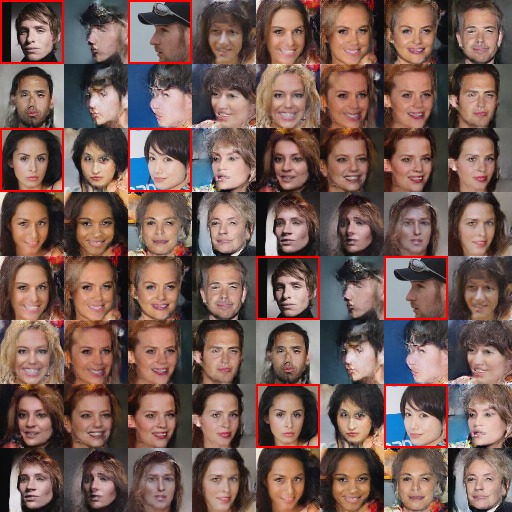}
    \caption{Manifold from a model trained on \emph{CelebA}. Images with red borders are taken from the training set, and define the manifold. The manifold was computed as described in Equation \ref{eq:manifold}, where the x-axis corresponds to $\phi$, and the y-axis to $\phi'$, and where $\phi, \phi' \in \{0, \frac{\pi}{4}, \cdots, \frac{7\pi}{4}\}$. }
\end{figure}

\begin{figure}[H]
\vspace{80pt}
    \centering \includegraphics[width=1.\textwidth]{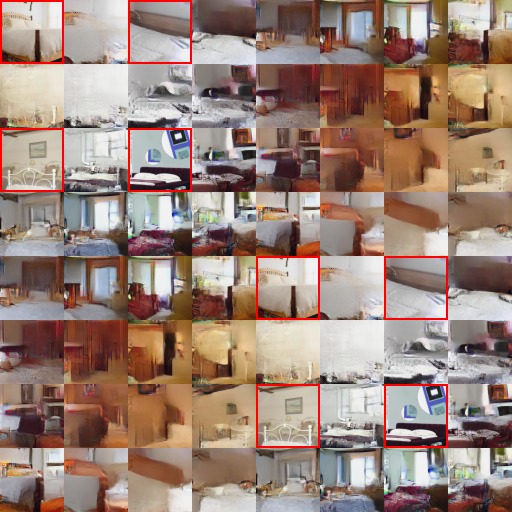}
    \caption{Manifold from a model trained on \emph{LSUN} (\emph{bedroom} category). Images with red borders are taken from the validation set, and define the manifold. The manifold was computed as described in Equation \ref{eq:manifold}, where the x-axis corresponds to $\phi$, and the y-axis to $\phi'$, and where $\phi, \phi' \in \{0, \frac{\pi}{4}, \cdots, \frac{7\pi}{4}\}$. }
\end{figure}

\begin{figure}[H]
\vspace{80pt}
    \centering \includegraphics[width=1.\textwidth]{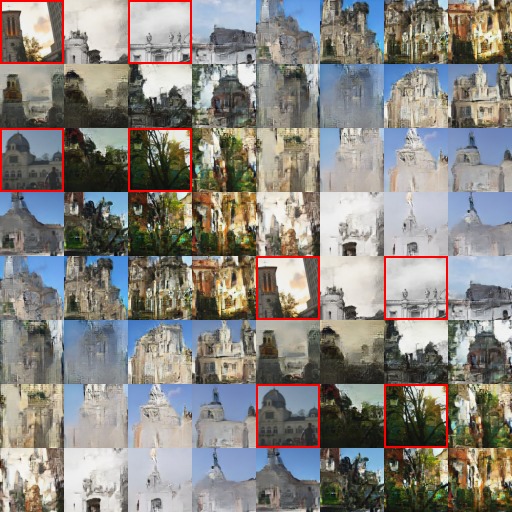}
    \caption{Manifold from a model trained on \emph{LSUN} (\emph{church outdoor} category). Images with red borders are taken from the validation set, and define the manifold. The manifold was computed as described in Equation \ref{eq:manifold}, where the x-axis corresponds to $\phi$, and the y-axis to $\phi'$, and where $\phi, \phi' \in \{0, \frac{\pi}{4}, \cdots, \frac{7\pi}{4}\}$. }
\end{figure}

\begin{figure}[H]
\vspace{30pt}
    \centering \includegraphics[width=1.\textwidth]{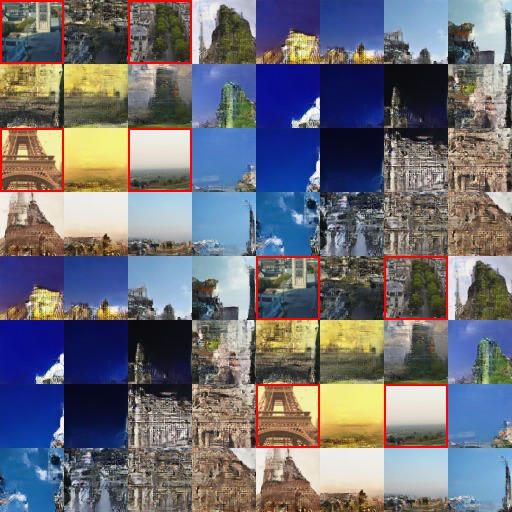}
    \caption{Manifold from a model trained on \emph{LSUN} (\emph{tower} category). Images with red borders are taken from the validation set, and define the manifold. The manifold was computed as described in Equation \ref{eq:manifold}, where the x-axis corresponds to $\phi$, and the y-axis to $\phi'$, and where $\phi, \phi' \in \{0, \frac{\pi}{4}, \cdots, \frac{7\pi}{4}\}$. }
\end{figure}

\section{Extrapolation}
Inspired by the texture generation work by \citep{DBLP:conf/nips/GatysEB15,theis2015generative} and extrapolation test with DCGAN \citep{DBLP:journals/corr/RadfordMC15}, we also evaluate the statistics captured by our model by generating images twice or ten times as large as present in the dataset.
As we can observe in the following figures, our model seems to successfully create a “texture” representation of the dataset while maintaining a spatial smoothness through the image. Our convolutional architecture is only aware of the position of considered pixel through edge effects in convolutions, therefore our model is similar to a stationary process. This also explains why these samples are more consistent in \emph{LSUN}, where the training data was obtained using random crops.
\newpage

\begin{figure}[H]
\vspace{0pt}
    \centering \subfigure[$\times 2$]{
      \includegraphics[width=.7\textwidth]{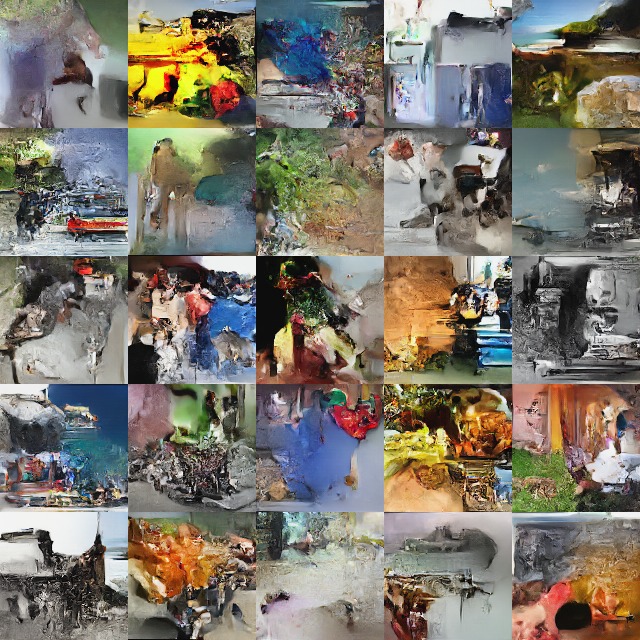}} \\
    \subfigure[$\times 10$]{
      \includegraphics[width=.7\textwidth]{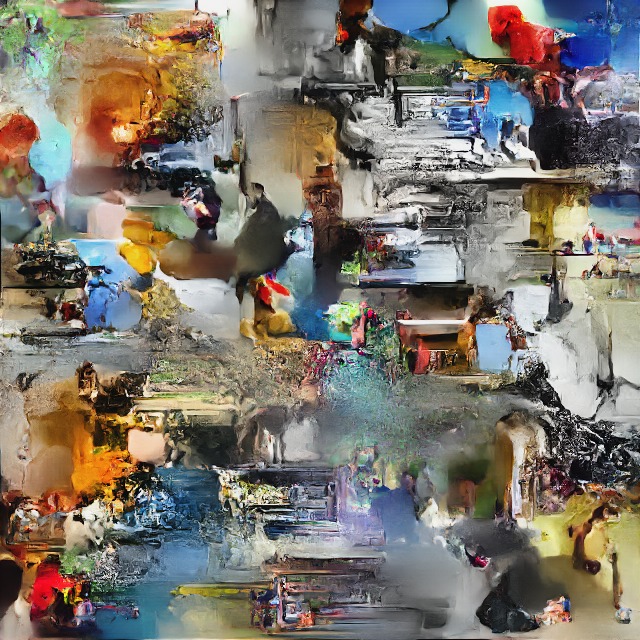}}
    \caption{We generate samples a factor bigger than the training set image size on \emph{Imagenet} ($64 \times 64$).}
\vspace{0pt}
\end{figure}

\begin{figure}[H]
\vspace{0pt}
    \centering \subfigure[$\times 2$]{
      \includegraphics[width=.7\textwidth]{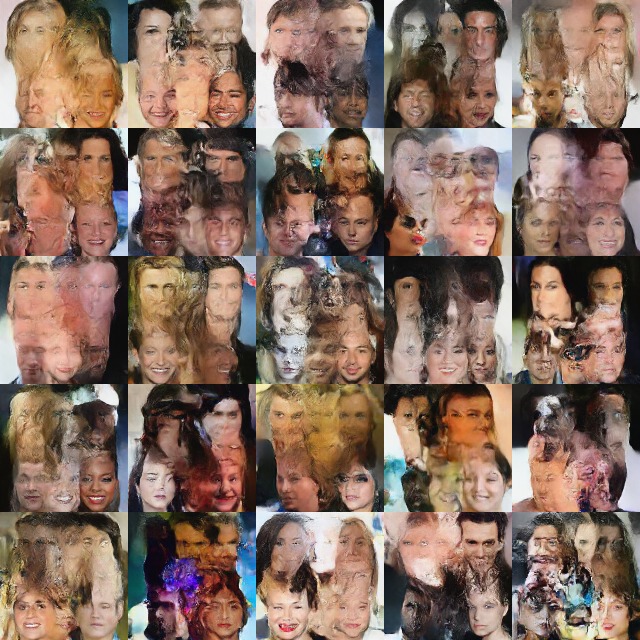}} \\
    \subfigure[$\times 10$]{
      \includegraphics[width=.7\textwidth]{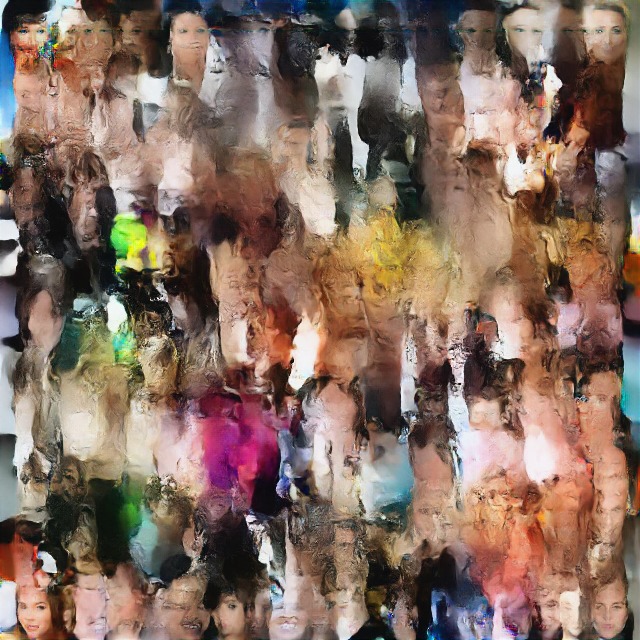}}
    \caption{We generate samples a factor bigger than the training set image size on \emph{CelebA}.}
\vspace{0pt}
\end{figure}

\begin{figure}[H]
\vspace{0pt}
    \centering \subfigure[$\times 2$]{
      \includegraphics[width=.7\textwidth]{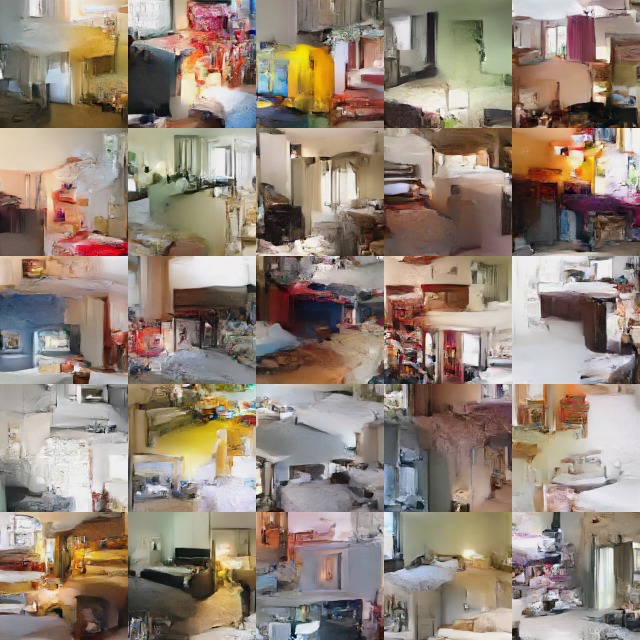}} \\
    \subfigure[$\times 10$]{
      \includegraphics[width=.7\textwidth]{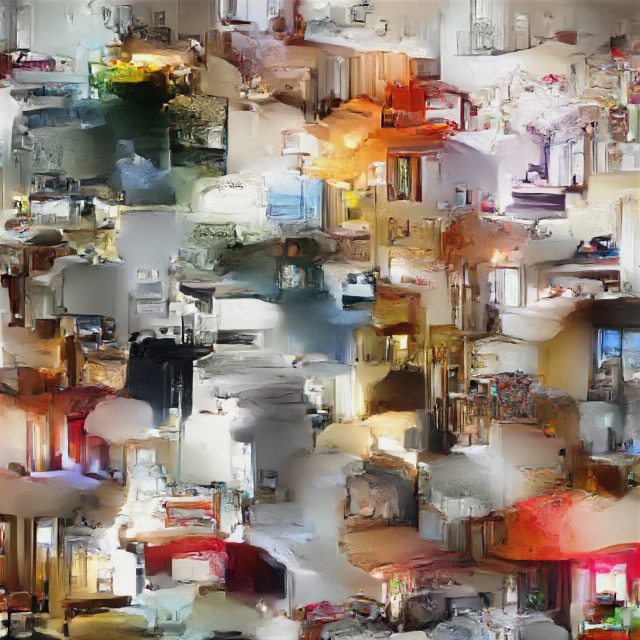}}
    \caption{We generate samples a factor bigger than the training set image size on \emph{LSUN} (\emph{bedroom} category).}
\vspace{0pt}
\end{figure}

\begin{figure}[H]
\vspace{0pt}
    \centering \subfigure[$\times 2$]{
      \includegraphics[width=.7\textwidth]{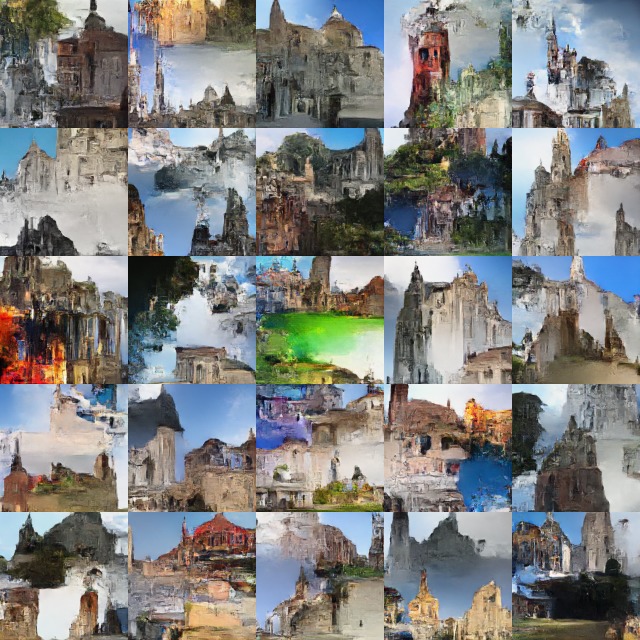}} \\
    \subfigure[$\times 10$]{
      \includegraphics[width=.7\textwidth]{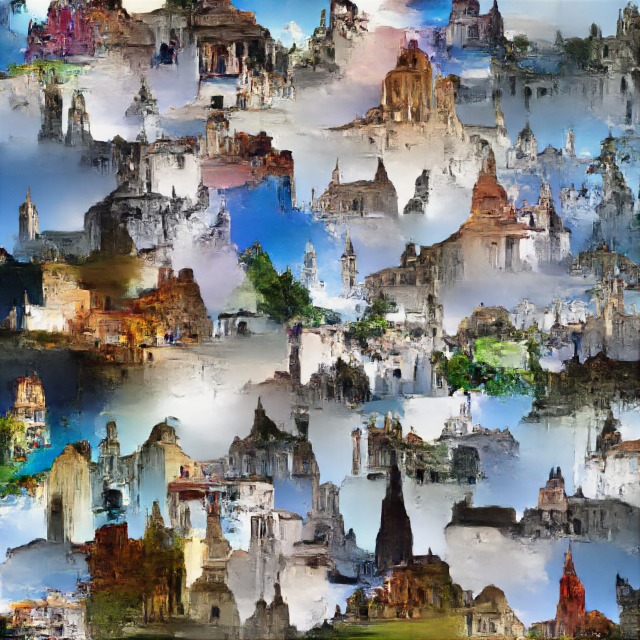}}
    \caption{We generate samples a factor bigger than the training set image size on \emph{LSUN} (\emph{church outdoor} category).}
\vspace{0pt}
\end{figure}

\begin{figure}[H]
\vspace{0pt}
    \centering \subfigure[$\times 2$]{
      \includegraphics[width=.7\textwidth]{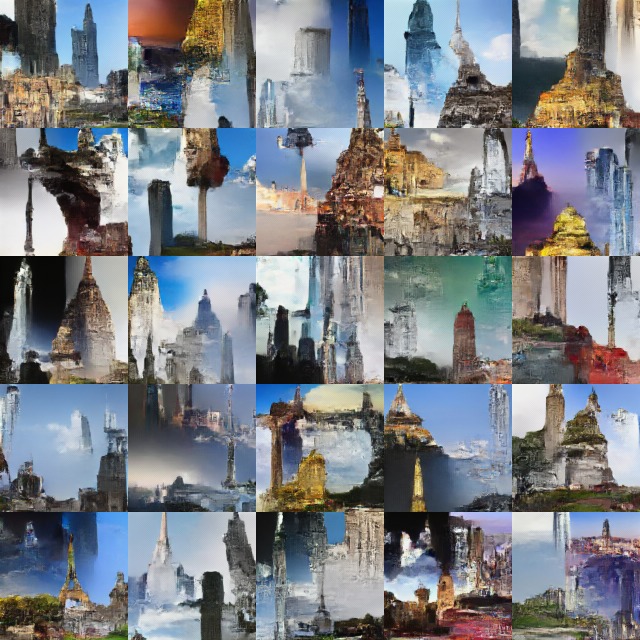}} \\
    \subfigure[$\times 10$]{
      \includegraphics[width=.7\textwidth]{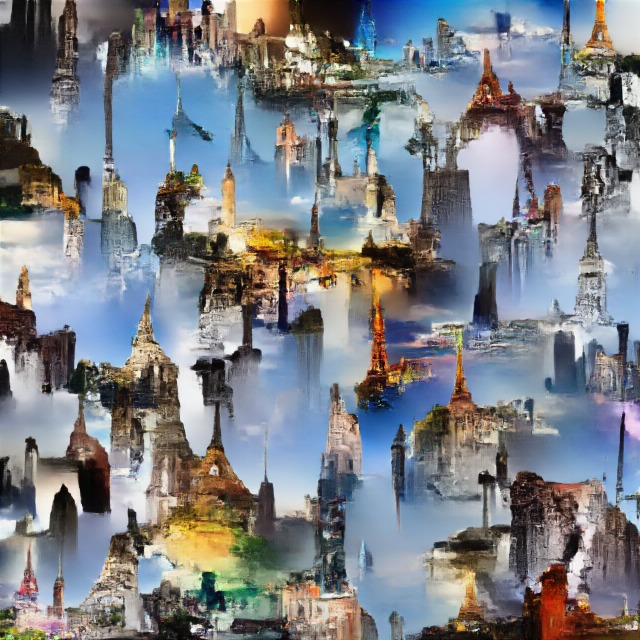}}
    \caption{We generate samples a factor bigger than the training set image size on \emph{LSUN} (\emph{tower} category).}
\vspace{0pt}
\end{figure}

\section{Latent variables semantic}
\label{app:latent}
As in \citep{gregor2016towards}, we further try to grasp the semantic of our learned layers latent variables by doing ablation tests. We infer the latent variables and resample the lowest levels of latent variables from a standard gaussian, increasing the highest level affected by this resampling. As we can see in the following figures, the semantic of our latent space seems to be more on a graphic level rather than higher level concept. Although the heavy use of convolution improves learning by exploiting image prior knowledge, it is also likely to be responsible for this limitation.

\begin{figure}[H]
    \centering \includegraphics[width=.5\textwidth]{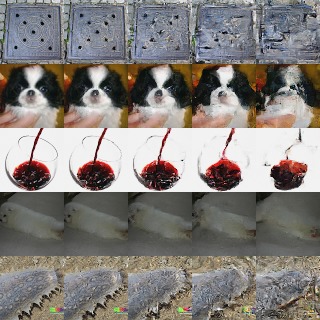}
    \caption{Conceptual compression from a model trained on \emph{Imagenet} ($64 \times 64$). The leftmost column represent the original image, the subsequent columns were obtained by storing higher level latent variables and resampling the others, storing less and less as we go right. From left to right: $100\%$, $50\%$, $25\%$, $12.5\%$ and $6.25\%$ of the latent variables are kept.}
\end{figure}

\begin{figure}[H]
    \centering \includegraphics[width=.5\textwidth]{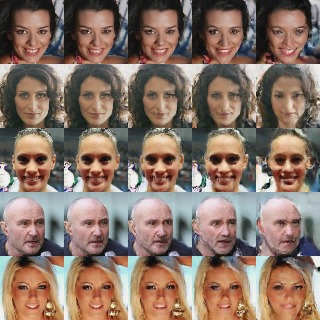}
    \caption{Conceptual compression from a model trained on \emph{CelebA}. The leftmost column represent the original image, the subsequent columns were obtained by storing higher level latent variables and resampling the others, storing less and less as we go right. From left to right: $100\%$, $50\%$, $25\%$, $12.5\%$ and $6.25\%$ of the latent variables are kept.}
\end{figure}

\begin{figure}[H]
    \centering \includegraphics[width=.5\textwidth]{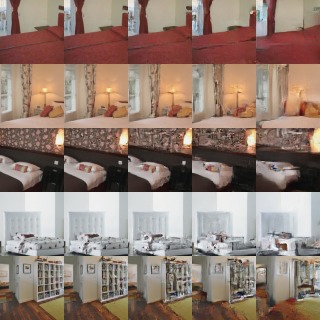}
    \caption{Conceptual compression from a model trained on \emph{LSUN} (\emph{bedroom} category). The leftmost column represent the original image, the subsequent columns were obtained by storing higher level latent variables and resampling the others, storing less and less as we go right. From left to right: $100\%$, $50\%$, $25\%$, $12.5\%$ and $6.25\%$ of the latent variables are kept.}
\end{figure}

\begin{figure}[H]
    \centering \includegraphics[width=.5\textwidth]{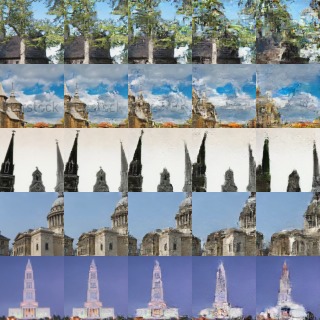}
    \caption{Conceptual compression from a model trained on \emph{LSUN} (\emph{church outdoor} category). The leftmost column represent the original image, the subsequent columns were obtained by storing higher level latent variables and resampling the others, storing less and less as we go right. From left to right: $100\%$, $50\%$, $25\%$, $12.5\%$ and $6.25\%$ of the latent variables are kept.}
\end{figure}

\begin{figure}[H]
    \centering \includegraphics[width=.5\textwidth]{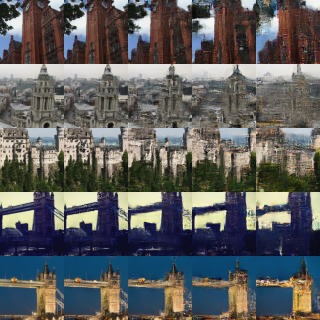}
    \caption{Conceptual compression from a model trained on \emph{LSUN} (\emph{tower} category). The leftmost column represent the original image, the subsequent columns were obtained by storing higher level latent variables and resampling the others, storing less and less as we go right. From left to right: $100\%$, $50\%$, $25\%$, $12.5\%$ and $6.25\%$ of the latent variables are kept.}
\end{figure}

\section{Batch normalization} \label{sec batch norm}
We further experimented with batch normalization by using a weighted average of a moving average of the layer statistics $\tilde{\mu}_{t}, \tilde{\sigma}^{2}_{t}$ and the current batch batch statistics $\hat{\mu}_{t}, \hat{\sigma}_{t}^{2}$,
\begin{align}
\tilde{\mu}_{t+1} &= \rho \tilde{\mu}_{t} + (1 - \rho) \hat{\mu}_{t} \\
\tilde{\sigma}^2_{t+1} &= \rho \tilde{\sigma}^2_{t} + (1 - \rho) \hat{\sigma}^2_{t},
\end{align}
where $\rho$ is the momentum.
When using $\tilde{\mu}_{t+1}, \tilde{\sigma}_{t+1}^{2}$, we only propagate gradient through the current batch statistics $\hat{\mu}_{t}, \hat{\sigma}_{t}^{2}$. We observe that using this lag helps the model train with very small minibatches.

We used batch normalization with a moving average for our results on CIFAR-10.

\section{Attribute change}\label{app attr change}
Additionally, we exploit the attribute information $y$ in CelebA to build a conditional model, i.e. the invertible function $f$ from image to latent variable uses the labels in $y$ to define its parameters. In order to observe the information stored in the latent variables, we choose to encode a batch of images $x$ with their original attribute $y$ and decode them using a new set of attributes $y'$, build by shuffling the original attributes inside the batch. We obtain the new images $x' = g\big(f(x; y); y'\big)$.

We observe that, although the faces are changed as to respect the new attributes, several properties remain unchanged like position and background.

\begin{figure}[H]
\vspace{80pt}
    \centering \includegraphics[width=1.\textwidth]{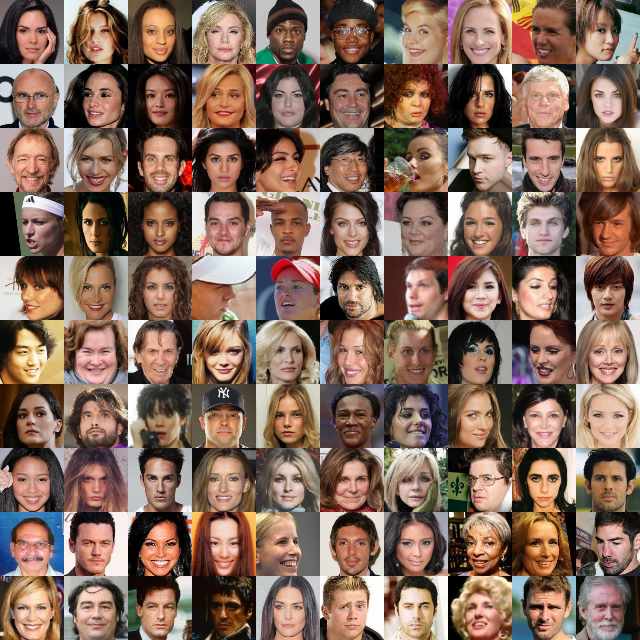}
    \caption{Examples $x$ from the \emph{CelebA} dataset.}
    \label{fig:transfer-original}
\end{figure}

\begin{figure}[H]
\vspace{80pt}
    \centering \includegraphics[width=1.\textwidth]{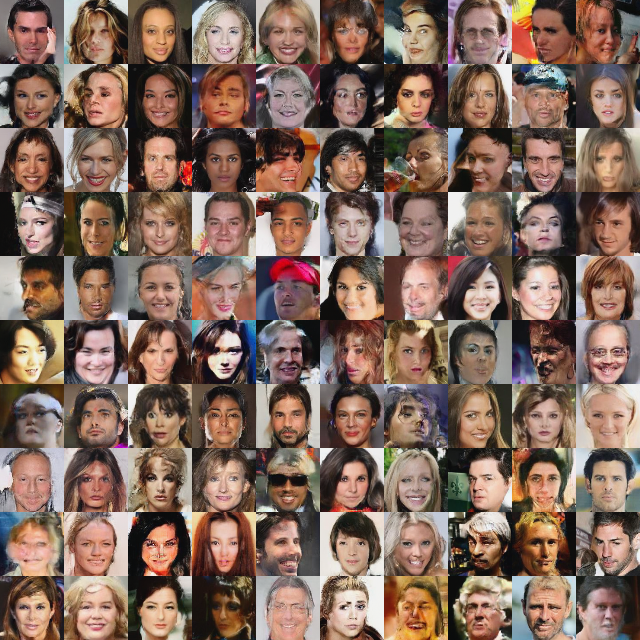}
    \caption{From a model trained on pairs of images and attributes from the \emph{CelebA} dataset, we encode a batch of images with their original attributes before decoding them with a new set of attributes. We notice that the new images often share similar characteristics with those in Fig \ref{fig:transfer-original}, including position and background.}
\end{figure}
\fi
\end{document}